\documentclass{article} % For LaTeX2e
\usepackage{nips15submit_e,times}
\usepackage{hyperref}
\usepackage{url}
\usepackage{graphicx}
\usepackage{subfig}
\usepackage{colortbl}

\usepackage[
backend=biber,
style=alphabetic,
sorting=ynt
]{biblatex}
\title{Routing and Placement of Macros using Deep Reinforcement Learning}

\author{
Mrinal Mathur\thanks{ Main Contribution} \\
\texttt{mmathur4@student.gsu.edu} \\
}

\nipsfinalcopy % Uncomment for camera-ready version
\begin{document}

\maketitle
\begin{abstract}
Chip placement has been one of the most time consuming task in any semi conductor area, Due to this negligence, many projects are pushed and chips availability in real markets get delayed. An engineer placing macros on a chip also needs to place it optimally to reduce the three important factors like power, performance and time. Looking at these prior problems we wanted to introduce a new method using Reinforcement Learning where we train the model to place the nodes of a chip netlist onto a chip canvas. We want to build a neural architecture that will accurately reward the agent across a wide variety of input netlist correctly.
\end{abstract}

\section{Introduction}

Due to increase in advancement of semiconductor industry, the world needs a special meet which can be solved using Artificial Intelligence. Since the decline of Moorse law, these industries are finding many other ways to keep the design. for example, we are entering the world of 3nm chips, the Moore's law is only applicable till 16-12nm. Due to this dramatic shift,the field of Machine learning is applicable in this domain. 

As the electronic devices evolved from few chips to millions of micro and nano chip placements incorporated in our electronic computing devices to improve on the performance and computation speed. Placement of these gates in these chips are very time consuming and complex. We present a learning based model to eradicate this complex and time consuming problem. Our main goal is to make the net list into a graph of standard cells (for example: AND gate, OR gate, etc) and place them on the chips by optimizing the power, performance and area of the chip while taking care of the constraints. \cite{https://doi.org/10.48550/arxiv.1708.04782}

\section{Related work}
Partition based methods: It works on basic principle of divide and conquer, i.e. split the netlist and chip canvas and keep splitting until it cannot be divided anymore. All these small units are then placed using optimal solvers. They have a limitation of compromise on quality with respect to routing congestion and also make it impossible to make changes in placement
 Stochastic/hill-climbing methods: Simulated annealing method applies random positioning of macros, measure objective function, if there is an improvement the change is applied else it will be applied with probability as temperature.This method generates high quality but has a drawback of slowness and inability to run parallel for large  and complex circuits.\\ 
 Analytic solvers: Forced directed methods and non-linear optimization approximations are part of advanced analytic solvers.Forced directed method works on the principle that nodes connected to each other are pulled closer and force disconnected nodes to pull away. this method is computationally efficient. Non-linear optimization uses math functions such as log-sum-exp, weighted average for wirelength and Gaussian and Helmholtz models for density. Latest being RePlAce, a mixed size placement technique that considers netlist like a charged particle with its area as electric charge, with penalty factor on local density for each bin size.\\
 All of the above have common downside of need for high computation power (GPUs) for complex circuit optimizations. DREAMPlace that originated from RePlAce algorithm, was proposed that resulted in speed of about 30x CPU.\\
 Graph neural network is used to represent each chip as a node and the wire connecting these chips as edges, which is then further modified to a .proto format for better communication within network during training.\\
 General approach is to learn from previous examples to speed up policy training for new unseen netlist. Joint learning based on reinforcement learning with optimization technique of gradient decent was proposed in one paper \cite{cheng2021on} \\
 As part of policy networks is concerned, recent trend proposed is on domain adaptation for combinatorial optimization, it improves on the performance and also reduce training time 8-fold. Another method is deep deterministic policy gradient that places logical cores and is rewarded or penalized based on latency among pipline stages. It works on the principle of policy to be parameterized through probability distribution. We came across proximal policy optimization algorithm that provide with option of multiple epochs of minibatch update capability, that is being used in our policy network implementation. \cite{https://doi.org/10.48550/arxiv.2004.10746}\\

\section{Datasets}
This is the most tricky part of this problem statement. We plan to take 28nm benchmark so that it is feasible for us to allow each moveable. We will be taking RISC-V architecture includes ISA level test. ARIANA RISC-V \cite{ariane} is a single issue, in-order CPU that implements the 64-bit RISC-V instruction set. We will be extracting macros from this benchmarks and for all configurable size, separate the TLBs for post processing. Dataset's Pre-processing  consists \cite{10.1137/S0895479895281484} of: 

\begin{enumerate}
    \item Mapping of standard cell libraries from netlist 
    \item placement information to be stored in DEF file 
    \item Actual Placement Layout (GDSII view) 
\end{enumerate}

These views are called standard cell physical design views and are used for timing analysis in Non-Linear Design Modeling (NLDM) architecture for making correct chip information in a library.

\section{Methods}
\subsection{Problem Statement}
We will target the chip placement optimization where we will be using the information from nodes in Netlist on the empty chip canvas because we want to optimize PPA : Power, Performance, Area. In the upcoming sections we will be explaining all the methods to implement the the optimization using a new method of policy learning. 

Despite rise in deep learning domains, Engineers are still iterating with the existing placement tools which takes a lot of time and efforts to correct even a single design. This complexity increases when the size of netlist increases and the scale of detail in grids on which these netlist needs to be placed, and high computation that takes weeks, Even after these problems the smaller the designs go, the more complex placements are. In order to address this challenge, we want to introduce a new method to pose chip placement that used Reinforcement Learning, where we train the agent using policy networks to optimize the placement of macros. we will be building a custom reward system that will approximately place the chips at each iteration. 

For our work we first cluster the standard cells by ignoring all the pin offsets of wire length and represent the placement canvas \cite{8418790}. We clustering the macros into a graph embedding and generate a reinforcement agent to learn to place macros on basis of policy in a blank grid instead of a occupied one. We require a mask generated by the post processing by the RL-agent. For the state space of 1000 cells, we have to place $10^10000$ states, There we need to first predict the representation of the given chip and then design an architecture to accurately predict the reward for the netlist and their placements for optimal PPA (Power, Performance and Area)

\subsection{Approach}

We will be using almost the same approach as the \cite{https://doi.org/10.48550/arxiv.2004.10746} but with a different approach for making it more reliable and accurate. The approach for placing macros using a Policy Network, we have considered learning with Proximal Policy Optimization Algorithms \cite{https://doi.org/10.48550/arxiv.1707.06347}. Policy Optimization Algorithm performs better than the state-of-art approaches while being very simple to implement and tune. The main challenge in using a Policy algorithm is that they are very sensitive to choice of placing the macros. Once we place all the macros we use an already famous algorithm called force-directed method \cite{https://doi.org/10.48550/arxiv.1012.4559} to place all the standard cells that we will be seeing in Figure1. We figured that we can re-formulate the same problem as a Markov decision process (MDP), Since the we are going to consider four elements while constructing this as a reinforcement learning problem. 

\begin{enumerate}
    \item States: All the netlist in the world along with all the parameters possible from placement tool (eg. Synopsis, Cadence). 
    \item Action: We have been placing current node onto the grid cell. 
    \item Transition: state transition define the probability distribution over the next states, so that we have information to go to state $s_{t+1}$ from state $s_{t}$ after updating the parameter
    \item Reward: we are using a custom reward functions which takes the negative weighted sum of wirelength, congestion and density. we will be taking an action in state: it will be zero for all the actions of until second last action where we update the reward every time.(Describe in detail in section [section]) 
\end{enumerate}

\subsection{Reinforcement Learning}
Recent breakthrough in computer science and natural processing has been relied on training which is very efficiently by using deep learning on all the training sets. Similar to other online approaches, we utilizes a technique known as experience replay where we store the agent's experience at each time-step, we will apply Q-learning updates by applying samples of experience. Learning directly from any samples is efficient, due to strong correlation between the samples. 

In the setting, at the initial state, $s_0$, we initial have an empty chip canvas and unplaced netlist. We will go till $T-1$ state with only our reward function and using our policy network (since $T$ is the equal total number of macros in netlist). At each step $t$, the agents in states ($s_t$), take an action ($a_t)$, arrives at a new state ($s_{t+1}$) and receives a reward ($r_t$) from environment.

In our Reinforcement learning environment, our agents learns to interact with the environment (chip canvas) by iterating over the number of states in given time stamp. After each timestamp, for each time $t$, the agent will have information related to state $s_t$, that which select which action $a_{t}$ is necessary for that timestamp from all the possible actions $A$ which has been learned due to policy $\Psi$, this $\Psi$ will map the states to actions. \cite{9256814}

We will keep on iterating until we reach the terminal state which restarts the whole process, we use an objective function for maximizing the return: 

\begin{equation}
    F(t) = \sum^N_{i=0}\Upsilon R_{t+N+1}
\end{equation}

where $\Upsilon$ is the factor for discarding the reward in future. 

The optimal policy will be the maximize the expected return or values. The values function $v_\Psi(s_t)$ is the expected return starting from state $s_t$ : 

\begin{equation}
    v_{\Psi}(s_t) = E[\sum^N_{i=0}\Upsilon R_{t+N+1} | s_t]
\end{equation}

\begin{figure}[h]
    \centering
    \includegraphics[width=75mm]{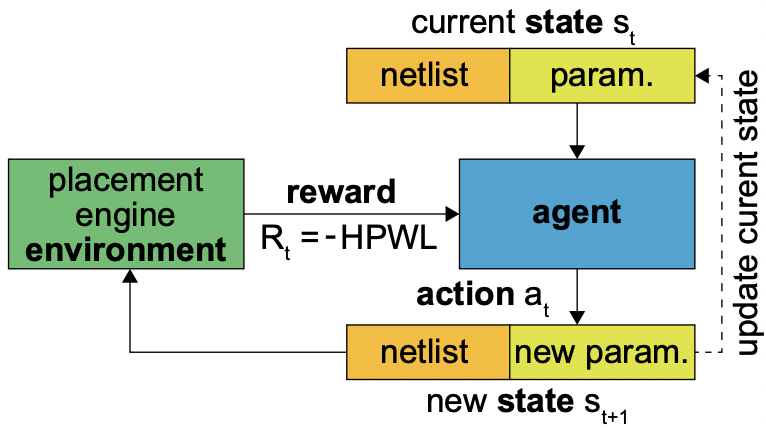}
    \caption{RL environment with proposed methods}
    \label{fig:RL_enviroment}
\end{figure}

In the case of chip placement design, as already mentioned, we have $2^{1000000}$ states. this makes it a $NP-hard$ problem to select the states for us initially, therefore we make use of bin packing formulation to solve it in polynomial time along with human expert representation.

we will explain how we represented states, action and rewards in subsequent sections.

\subsubsection{State}
We used metrics from each netlist needs to be placed and defined states as our parameter from Synopsis tool which is used for placement in industries still. Netlists parameters includes metadata information that is necessary for placement, this information includes: number of cells, area knowledge of each floorplaninig, etc. We took all the information for nodes/macros as edges and the wirelength connecting them as edges and converted this information into a graph representation that has topological features. Since we need to transfer this knowledge across many other macros so that it can generalize among unseen netlists. 

We used SAGE \cite{sagemath} to convert our macros into graph format by using topological characteristics like connections and spectral using: Strongly Connected Components, Cliques and k$-$graph coloring. 

The following features 
We used logic levels, that is to find the maximum distance between two flip flops
\begin{equation}
    Logic Levels = max_{x, y \in FFs} d(x, y)
\end{equation}

Along with this, we needed the information of closeness of features, in NLP we can formulate this by using embeddings like Word2Vec but here we used Rich Club Coefficients which tells how we do two nodes know each other and the measure of fractal coloring which gives information on cliques closeness and it's neighbors, this is given as: 

\begin{equation}
    CC = \frac{1}{V} \sum_{i\in V} \frac{|e_{jk} : v_j, v_k \in Neighbors(i), e_{jk} \in E|}{deg(i) (deg(i) - 1}
\end{equation} 

Using this we extract spectral characteristics by restarting Arnoldi method \cite{10.1137/S0895479895281484} , we extract Laplacian matrix of our proposed graph of netlists G the Fiedler value (second smallest eigenvalue) which is deeply connected to the connectivity properties of graph G, as well as spectral radius (largest eigenvalue) related to G.

These features are important for extracting information like congestion (which is considered during placement refinement) and logic levels extracts the longest logic path which solves the problem of meeting timing since finding longest path is also a NP-hard problem. \cite{10.1145/1055137.1055186}

\begin{figure}[h]
    \centering
    \includegraphics[width=90mm]{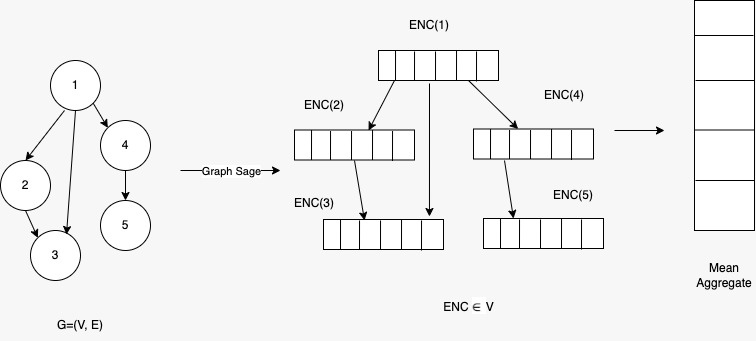}
    \caption{graph representation used for GNN}
    \label{fig:graph}
\end{figure}

\subsubsection{Implementing Graph Neural Network}
We added node features including encoded grate type, fanout area, degree. We generated embedding using GraphSAGE \cite{https://doi.org/10.48550/arxiv.1706.02216}  with aggregation of convolution, dropout and output size to be 32. These embedding $(ENC(v))$ are really important factors which basically converts all the physical information into a vector representation that attractively propagates information from node to its neighbors.

We implement our custom policy update factor which is compatible with Stochastic Gradient Descent \cite{doi:10.1137/100802001}  and simplifies the algorithm by removing any KL divergence penalty that needs to make adaptive update. We use Proximal Policy  Optimization \cite{https://doi.org/10.48550/arxiv.1706.02216} \ref{fig:graph} which enables use to learn the policies before we plug it to RL environment. The T-SNE projection into two dimension of the graph features is shown in \ref{fig:tsne}, We can see that all the features are far apart, which means that combination of our learned graph network distinguishes well on each netlist. (Information on our netlist features is given in Appendix)

\begin{figure}[h]
    \centering
    \includegraphics[width=80mm]{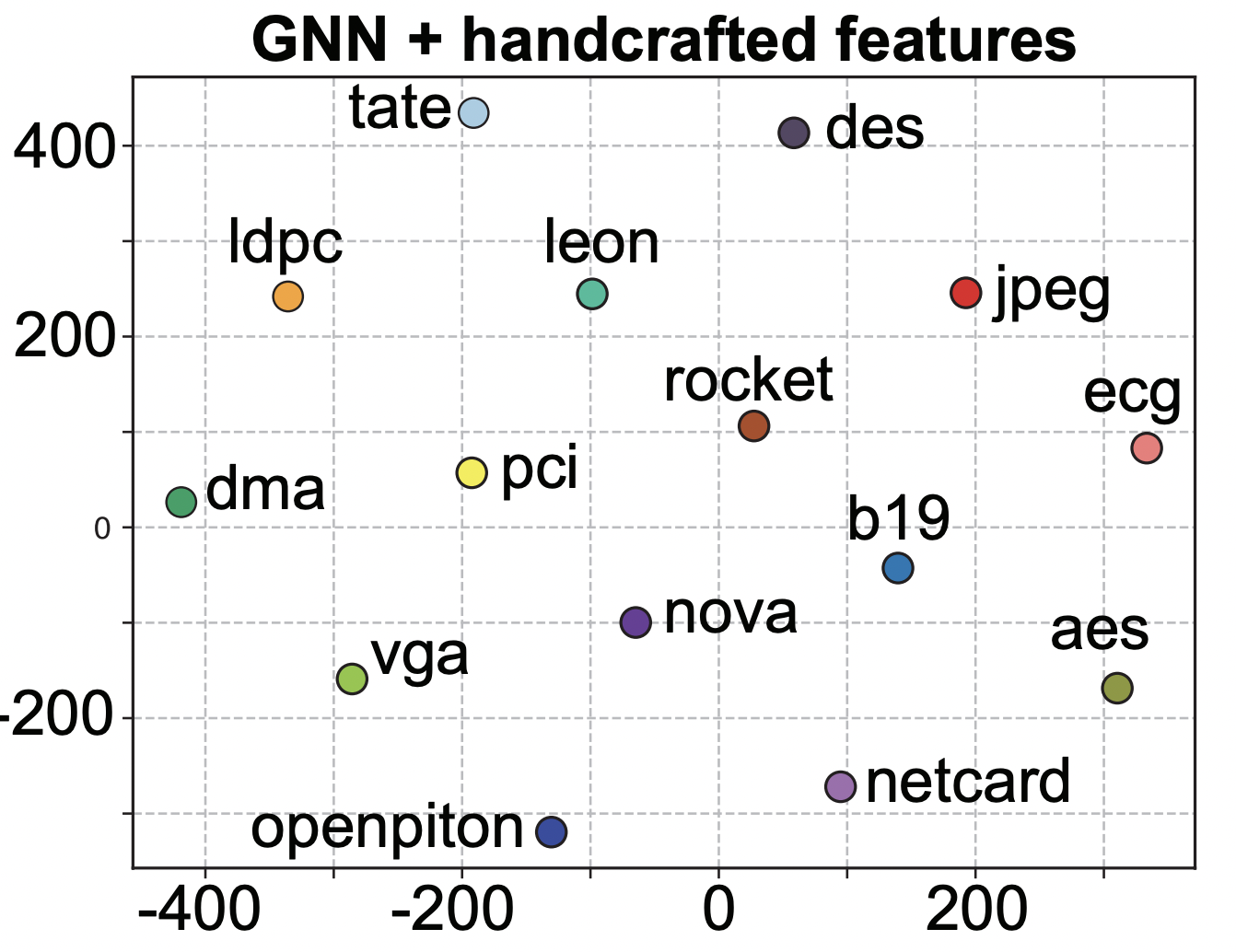}
    \caption{T-SNE representation of our Graph Neural Network features combined}
    \label{fig:tsne}
\end{figure}

\subsubsection{Action}

we want our actions to be in such a way that the placement is unique and should not be repeated, this is for the sole reason that we don't want any macros to be placed on one on another. So we use the given state action pair and using our Proximal Policy Optimization Algorithms \cite{https://doi.org/10.48550/arxiv.1707.06347}, we get the unique state $s_{t+1}$

Starting from state $s_0$ and using the learned policy $\Psi$, we can compute the state $s_n$ directly without performing any placement. We designed our action space such that it will first look at the functionality of the macros first and then the placement space later, this allowed us to reduce the search space and perform placement faster.

\subsubsection{Reward}
We define our reward function for RL agent for various netlist that includes different wirelength and congestion values. Our first proposed reward function was, we used half-perimeter wirelength 

\begin{equation} \label{eq:reward}
    R_t = - HPWL - congestion_{t}
\end{equation}

But \ref{eq:reward} was slower and was taking more than required time, So to help convergence we normalized the reward function by introducing the reward for each time. 

\begin{equation} \label{eq:reward_2}
    R_t  = \frac{- HPWL_t - congestion_{t}}{HPWL_t}
\end{equation}

Using \ref{eq:reward_2}, we can also adjust the parameter for adjusting to target PPA for any reward function for any technology.

\subsection{Proposed Architecture}
We are using Policy network with already learned policy to predict the placement of the macros, so our first architecture is going to be a Graph Neural network with  

\begin{figure}[h]
    \centering
    \includegraphics[width=100mm]{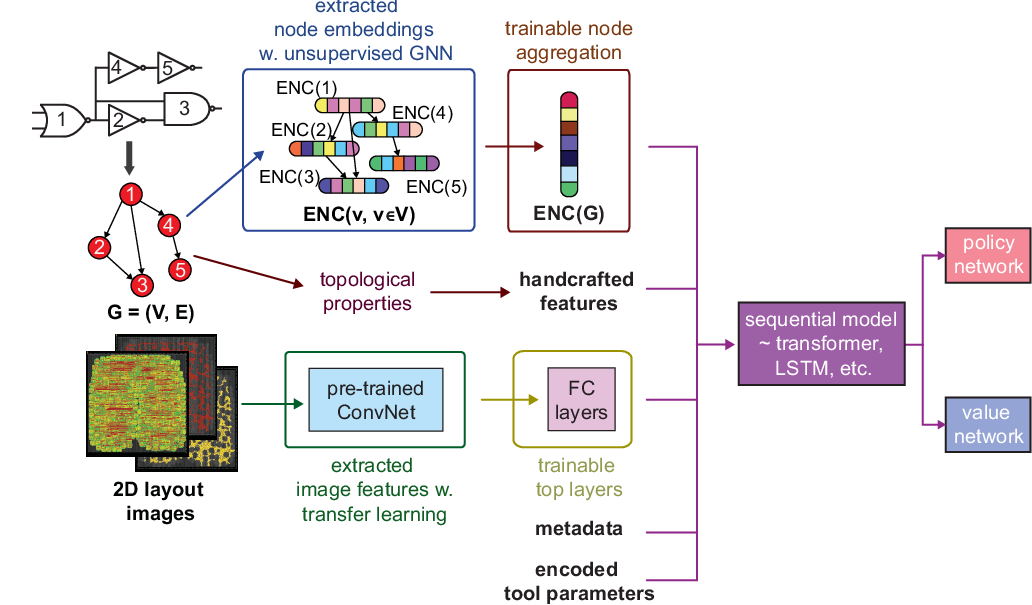}
    \caption{Proposed architecture }
    \label{fig:tsne}
\end{figure}

We have already mentioned how we trained our Graph Neural Network, The supervised model is trained to minimize the weighted RMS score to first learn all the policies and then save them in the embedding layer we dont have to learn them again and again. After transfering our information into Latent space, we apply policy-value network to place our macros with the appropirate reward function.

In this approach we have only done zero-shot learning since to pretrain the network it takes significant computation but we will show results for pretrained done by Qualcomm team.

In our Policy Network, we compute "How well is our placement setting getting update" and Value Network computes "How good is our current settings". we share parameteres with both the networks simultaneously to predict and inform one another. These parameters are adjusted by the gradients over the sum of losses of policy and value networks, we also added regularization term to avoid the risk of overfitting. The equation of flow of gradients is as follows: 

\begin{equation}
    \underbrace{(G_t^{(n)} - v_w(s_t))\nabla_{\theta}\log\Psi_{\theta}(a_t|s_t)}_{policy gradient} +\underbrace{\beta(G_t^{(n)} - v_w(s_t))\nabla_{w}v_w(s_t)}_{valueestimated gradient}+\underbrace{\eta \sum \Phi_{\theta}(a_{\theta}|s_t)\log\Psi_{\theta}(a|s_t)}_{entropy regularisation}
\end{equation}

We added entropy regularisation to push the entropy up to encourage exploration and $\beta$ and $\nabla$ are hyper-parameters to balance loss components.

The input for Policy and Value networks are embeddings of macros or netlist graph, the id of current node and meta data of netlists to generate the possible distribution over actions and pass it to value network to predict the next state from the original state. We use transformers [cite hugging face] for faster calculation of the states and better results on predictions. 

\section{Results}
To train and test our approach we selected one RISC-V single core (ARIANE) and two RISC-V multi-core tested by Qualcomm. These benchmarks each have more than 60000(approximately) macros for each core. we took 80\% for traning and 10\% each for testing and validation. 
We first synthesised the netlist using Synopsis Design Compiler for TSMC 29nm technology node and Placement was done by Cadence Innovus 18.1. All the information for floorplanning was discussed and closed at 1 aspect ration with fixed clock frequency. 

The First memory macro were placed by hand by Enginees, with maximum density which was calculated by lower bound of total cell area.

For RL enviroment, we used OpenAI Gym Interface and TensorFlow RL library to implement our RL agent. We used nitro cluster for training and preprocessing for the macros where we initally started training for 100 epochs (~16,0000 placements ) which took 100 hours straight. This time was taken to update the parameters and spend time learning the policies and placement he macros with updated weights.

we evaluated our results with RePLACE P$\&$R system as well as with Human Engineer to check how well our methods have worked as shown in Table \ref{tab:result-table}

We will show results divided in these area: 

\begin{enumerate}
    \item Timing analysis which captures Total power(W), Total Area($\mu m^2$) and Total Performance(ps)
    \item we will show performance difference between graph placement and Hypergraph placement 
    \item the effect of dataset in Policy Network for placement
\end{enumerate}

\begin{table}
\centering
\begin{tabular}{|>{}lllll|}
    \hline
    Methods& Timinig &Area($\mu$ $m^2$)&Power(W)&Wirelength(m)\\
    \rowcolor{gray}Manual&233.2&169341&3.24&52.14\\
    \rowcolor{gray}RePLACE&87.9&168724&3.73&51.11\\
    \rowcolor{lightgray}Ours&52&168645&3.23&51.09\\
\end{tabular}
\caption{\label{tab:table-name}Results Depecting our results with respect to other methods.}
\label{tab:result-table}
\end{table}

\begin{figure}%
    \centering
    \subfloat[\centering Our Methods]{{\includegraphics[width=5cm]{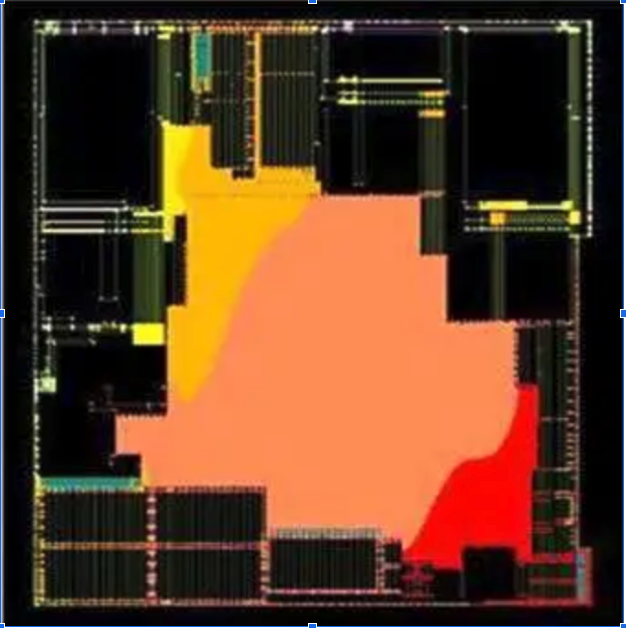} }}%
    \qquad
    \subfloat[\centering  Engineer work in 4 days]{{\includegraphics[width=5cm]{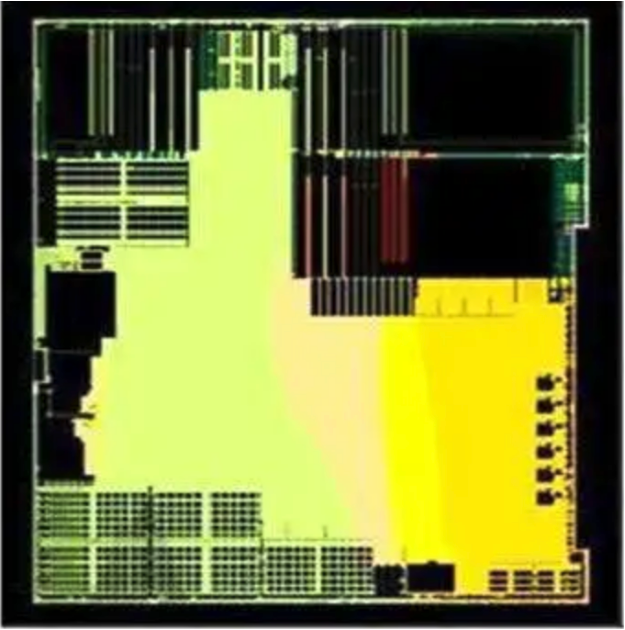} }}%
    \caption{Results on Qualcomm FloorPlan}%
    \label{fig:Results1}%
\end{figure}

We also show two technology nodes that are ran by Qualcomm on Confidential Floorplan node as shown in Figure \ref{fig:Results1}

\begin{figure}[h]
    \centering
    \includegraphics[width=90mm]{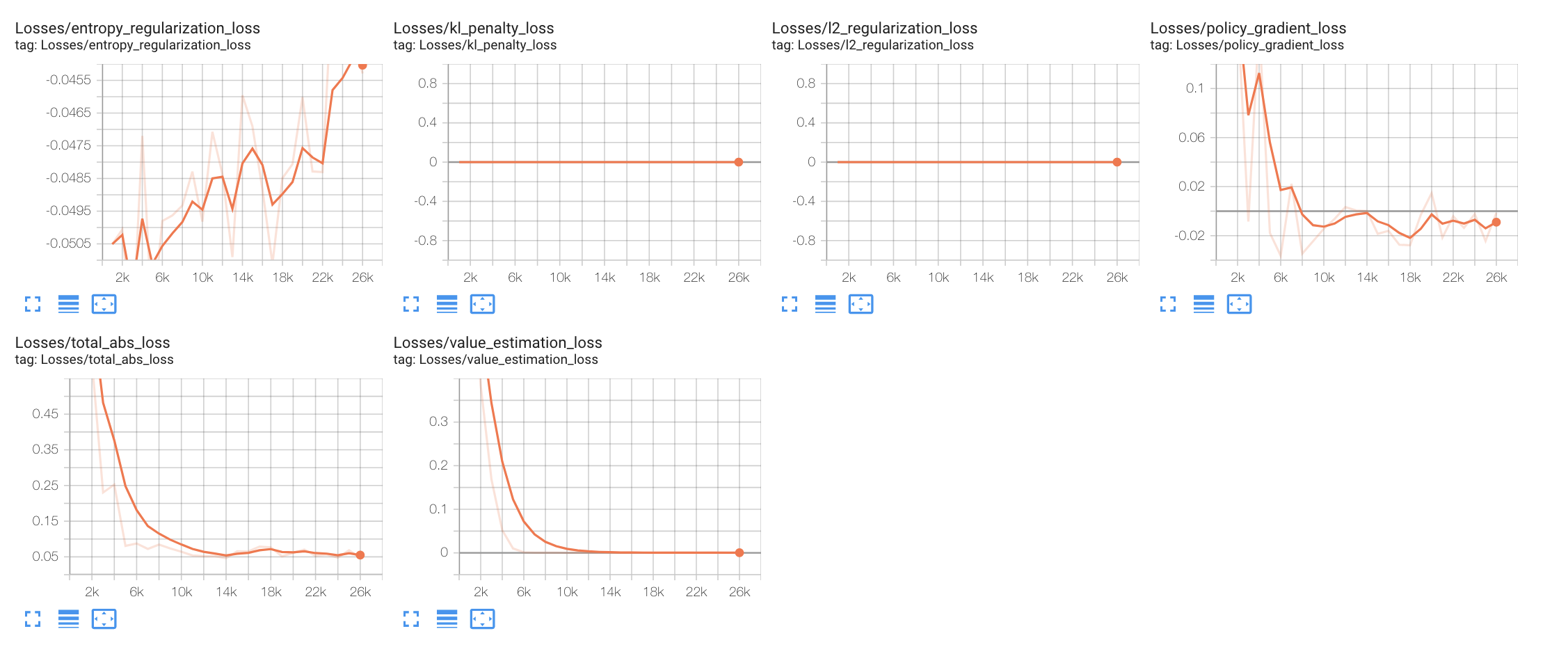}
    \caption{Metrics plot}
    \label{fig:metric}
\end{figure}

\begin{figure}[h]
    \centering
    \includegraphics[width=90mm]{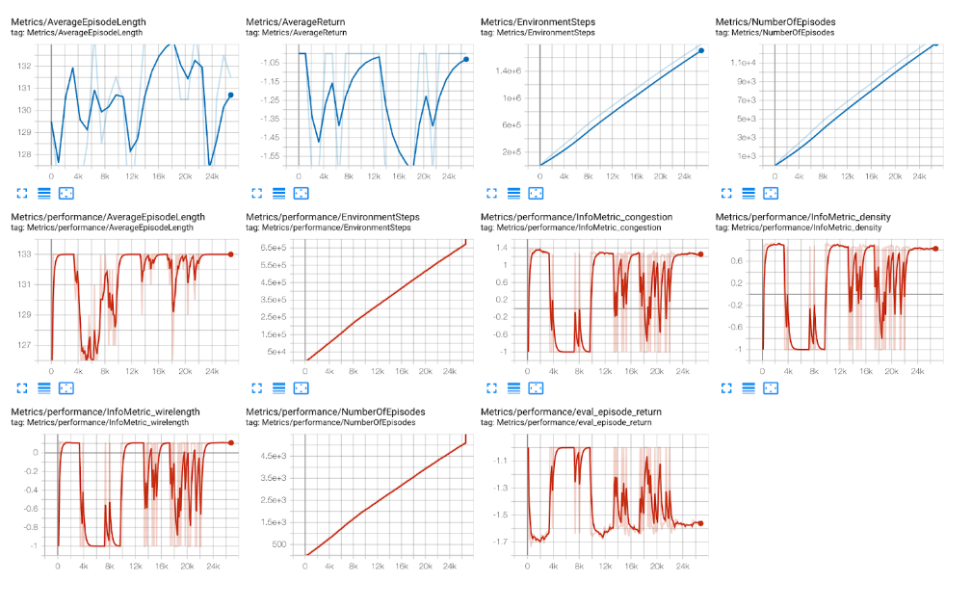}
    \caption{Losses plot}
    \label{fig:loss}
\end{figure}

\begin{figure}[h]
    \centering
    \includegraphics[width=90mm]{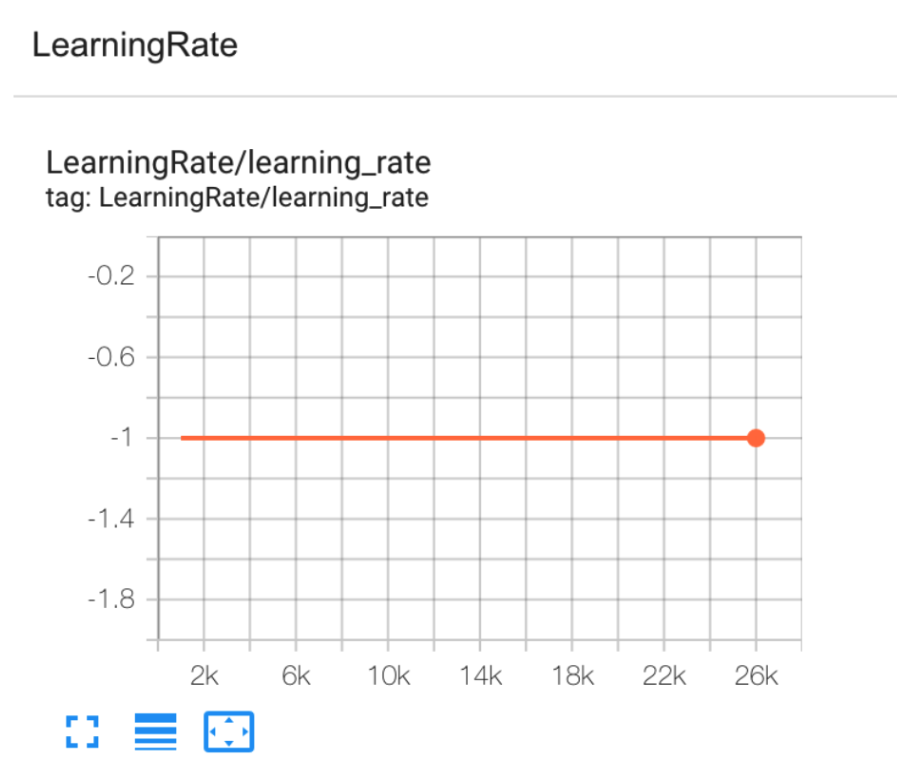}
    \caption{Learning rate plot}
    \label{fig:lr}
\end{figure}

Figure \ref{fig:metric} , \ref{fig:loss}, \ref{fig:lr} shows the convergence and learning plots for training from scract.  we havent done this on Pretraining so If time permits and we have enough resources, we want to training a pre-trained model to a new netlist with 1-shot, 3-shot, etc and fine-tuning parameteres

\section{Conclusion}
Placement based tasks are very time consuming and we show in our work that by targeting complex and impactful problem across industry we will be able to solve the problem. We propose a new RL-based approach to place the macros faster and in efficient manner to optimize PPA values. We show that we have outperform state-of-the-art baselines and generated placements with better results. Our results show that our agent generalized well and reduces the wirelength compared to EDA tools without any additional training cost.

\section{Acknowledgement}
We sincerely thank our Professor, Dr. Jonathan Shihao Ji, for his support in providing us with required GPU computation, guidance and teachings that led us undertake this project and gain more practical experience on understanding the Deep learning concepts that were taught in class.

We would like to thank Google research team, Azalia Mirhoseini, Ebrahim Songhori, Jeff Dean for their guidance.

Also, Ronita Mitra from Qualcomm for helping us with the inital Placement on EDA tools and visualizations.

\medskip

%%%%%%%%%%%%%%%%%%%%%%%%%%%%%%%%%%%%%%%%%%%%%%%%%%%%%%%%%%%%%%%%%%%%%%%%%%%%%%%%%%%%%%%

\small

\printbibliography
\end{document}